\documentclass[journal]{IEEEtran}
\usepackage{cite}
\usepackage{graphicx}
\usepackage{amsmath}
\usepackage{amssymb}
\usepackage{enumitem}
\usepackage{booktabs}
\usepackage{multirow}
\title{Lightweight ML-Based Automatic Sleep Staging Framework with Constrained CNN and Mamba for Small-Sample EEG Datasets}
\author{
    \IEEEauthorblockN{Zihao Wei\textsuperscript{1}, $^*$Yulin Gong\textsuperscript{1}, Yudan Lv\textsuperscript{2}}

    \IEEEauthorblockA{
        \textsuperscript{1}\mbox{School of Electronic Information Engineering,}\\
        \mbox{Changchun University of Science and Technology,}\\
        \mbox{Changchun 130022, Jilin Province, China}
    }

    \IEEEauthorblockA{
        \textsuperscript{2}\mbox{Jilin University,}\\
        \mbox{Changchun 130012, Jilin Province, China}
    }

    \IEEEauthorblockA{
        \mbox{$^*$Corresponding author: Yulin Gong, Email: gongyulin@cust.edu.cn}
    }
}

\begin{document}

\maketitle

\begin{abstract}
Automatic sleep staging is a key technology for precise diagnosis and treatment of sleep disorders as well as long-term home sleep monitoring. Portable electroencephalogram (EEG) devices have become the focus of research due to their convenience in data collection. However, current methods still face three major challenges: large parameter sizes that easily lead to overfitting on small datasets, low accuracy in classifying difficult stages such as N1 and REM, unclear optimal training dataset size, and difficulty in deployment. This paper proposes GamSleepNet, a lightweight and low-latency automatic sleep staging framework for single-channel EEG. The framework features the FEB module, which combines improved Gabor kernels with learnable filters for feature extraction, uses the Mamba architecture to build a temporal classification network, introduces a novel contrastive loss and a two-stage training strategy, and experimentally validates the optimal dataset size for single-channel EEG sleep staging models. On the Sleepedf dataset, this model achieves an overall accuracy of 87.86 percent with only 30.86 thousand parameters, with all metrics reaching SOTA levels and significantly improving the identification accuracy of challenging sleep stages.
\end{abstract}


\section{Introduction}
Sleep staging is a key process in the precise diagnosis and treatment of sleep disorders, with polysomnography (PSG) recognized as the clinical gold standard for sleep staging. According to the guidelines of the American Academy of Sleep Medicine (AASM)\cite{1,2}, the sleep specialists are required to divide PSG data into 30-second segments (epochs) and classify them into five stages: Wake, non-rapid eye movement stages N1, N2, N3, and rapid eye movement stage REM.

Traditional manual sleep scoring is a tedious and time-consuming task\cite{3} that is difficult to meet the needs of large-scale clinical screening and long-term home sleep monitoring. As a result, automatic sleep staging algorithms based on portable devices have become important technological directions for home sleep monitoring due to their convenience of data collection and the portability of equipments.

Currently, automatic sleep staging methods are typically based on convolutional neural networks(CNN)\cite{4,5,6}, recurrent neural networks(RNN)\cite{7}, fully convolutional networks(FCN)\cite{8,9}, Transformer models\cite{6,10,11}, and hybrid architectures\cite{12,13,14,15,16}. The emergence of these models has enabled automatic sleep staging to reach a level comparable to that of sleep specialists\cite{17}.

However, the methods menthioned above still face three major challenges: First, the number of model parameters and computational requirements are relatively high\cite{11,16,18}, making it very easy to overfit when training on small datasets commonly used for deploying portable devices. Second, EEG patterns across different sleep stages are highly similar, which can easily lead to classification confusion and consistently low accuracy for challenging stages such as N1 and REM\cite{19,20}. Third, model performance varies significantly across datasets of different sizes\cite{11,16,21,22,23}, and the optimal amount of training data required remains unclear.

To address these above issues, this study proposes a lightweight ML-based framework with constrained CNN and Mamba for small-sample EEG datasets. In this study, a learnable feature extraction block (FEB) is designed using an improved Gabor kernel function\cite{23,24,25,26} and incorporates a learnable filter\cite{7,15} branch, enabling efficient extraction of multi-scale time-frequency features from EEG signals with a minimal number of parameters firstly. Secondly, a temporal classification network is constructed using the Mamba architecture\cite{27,28,29}, which replaces traditional Transformer\cite{28,31}, LSTM\cite{32}, GRU\cite{33} and RNN\cite{30}, achieving long-range temporal modeling with linear complexity and further reducing computational cost. In addition, a novel contrastive loss computational strategy is introduced to align feature distributions across models, enhance the model's ability to distinguish challenging sleep stages, and alleviate class imbalance. Finally, systematic experiments determine the optimal dataset size for training single-channel EEG sleep staging models.

The main contributions of this study are as follows:
\begin{enumerate}[leftmargin=*]
    \item A novel network architecture is proposed based on Gabor kernel functions and the Mamba architecture, enabling automatic sleep staging with a small number of parameters and high computational efficiency.
    \item A novel contrastive loss calculation method is designed, which computes contrastive loss between the encoded outputs of the target model and state-of-the-art(SOTA) models. Combined with a two-step training algorithm, this approach improves the accuracy for challenging sleep stages.
    \item The optimal dataset size was validated through experiments conducted on public Sleepedf and private datasets.
\end{enumerate}
\section{Method}

\subsection{Overall Framework}
\begin{figure}[!t]
\centering
\includegraphics[width=\linewidth]{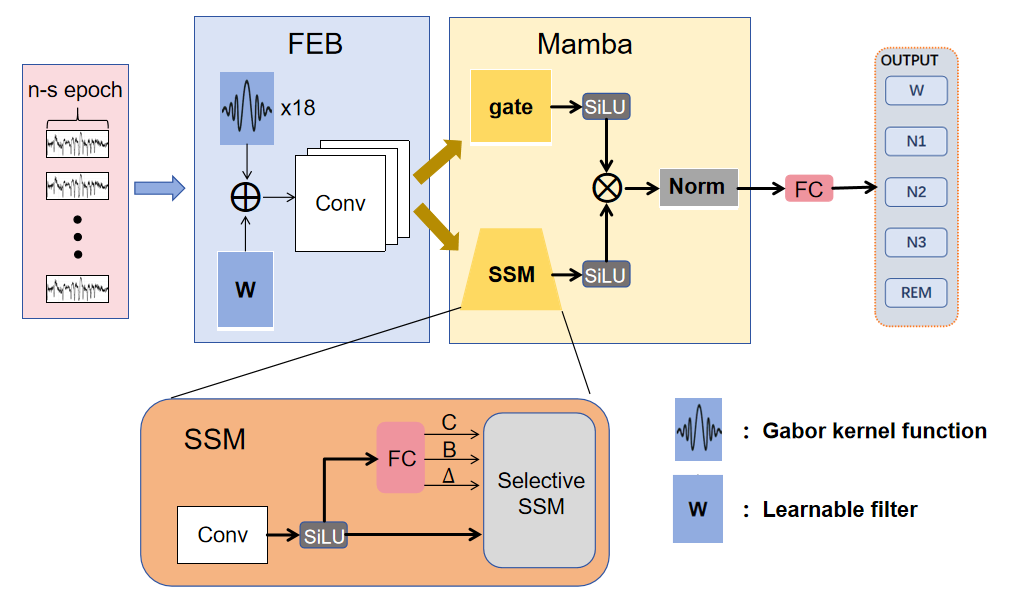}
\caption{Architecture of the proposed model, which consists of two components: the FEB learnable feature extraction module and the Mamba temporal classification network.}
\label{fig:framework}
\end{figure}

This paper proposes a portable EEG automatic sleep staging model(GamSleepNet). As shown in Fig.~\ref{fig:framework}, the overall framework consists of two core components: the FEB learnable feature extraction module and the Mamba temporal classification network \cite{27,28,29}. In addition, a two-stage training strategy and a novel contrastive loss function are designed to further optimize the model's performance.

The FEB module is based on the improved Gabor kernel functions and integrates a learnable filter branch with one-dimensional convolution to efficiently extract features from EEG signals using fewer parameters. The classification network is centered around the Mamba module, which models temporal context through a gating branch and an SSM branch, and finally outputs sleep staging predictions through a fully connected layer.

The model uses a two-stage training approach. First, the parameters of the FEB module are optimized. Then, the FEB is frozen and the proposed contrastive loss is introduced to jointly optimize the classification network, effectively alleviating the problem of class imbalance.

\subsection{FEB}
To achieve more efficient feature extraction with fewer parameters, this paper draws on the Gabor kernel functions proposed in \cite{23,24,25,26} and additionally introduces two parameters, offset ($u$) and phase ($P$). Based on this, a learnable feature extraction network called FEB is proposed.

\begin{equation}
G(t^*) = e^{-\pi \frac{(t^* - u)^2}{|\sigma|}} \cos\left(2\pi f t^* + P\right)
\label{eq:improved_gabor}
\end{equation}

The Gabor kernel function contains four trainable parameters where $u$ represents the temporal offset, $\sigma$ is the standard deviation, $f$ is the central frequency of the kernel, and $P$ is the phase. The waveform characteristics of the Gabor kernel are similar to those of EEG signals, which enables it to mimic EEG waveforms through matching. This matching is achieved by using $i$ Gabor kernels as convolutional kernels, which are cross-correlated with the input signal $X(t)$ to produce $i$ output layers.

\begin{equation}
GL^i(t) = G^i(t^*) \star X(t)
\label{eq:gabor_cross_correlation}
\end{equation}

\begin{equation}
Y = W \times X(t)
\label{eq:learnable_filter}
\end{equation}

Since standard Gabor kernel functions have limited ability to match EEG waveforms, a learnable filter is introduced during the cross-correlation operation. The output $Y$ of the learnable filter is added to $GL^i(t)$, and the sum is passed through a one-dimensional convolutional layer to obtain the final output of the FEB layer.

\subsection{Classifier Network}
The classification network consists of two parts, Mamba and fully connected layers, which are used to model the temporal sequence of the extracted features and output the final prediction results.

The Mamba module is composed of multiple Mamba blocks. Each block first normalizes the input $x$, then uses a linear projection to divide the normalized features into a gating branch $x_z$ and an SSM branch $x_u$. The SSM branch passes through causal convolution and SiLU activation to obtain $x_{\text{conv},t}$, which is then sent into the core selective SSM unit. The computational logic of this unit is as follows:

\begin{equation}
h_t = \overline{A}_t h_{t-1} + \overline{B}_t x_{\text{conv},t}
\label{eq:ssm_state}
\end{equation}

\begin{equation}
y_t = C_t \cdot h_t
\label{eq:ssm_output}
\end{equation}

In the above equations, $t$ represents the time step. $\overline{A} = \exp(\Delta A)$, $\overline{B} = \Delta B$, and $\Delta = \text{softplus}(\text{Linear}(x_{\text{conv},t}) + \text{bias})$. $B = \text{Linear}(x_{\text{conv},t})$ and $C_k = \text{Linear}(x_{\text{conv},t})$. $\text{Linear}(\cdot)$ denotes a linear projection layer, and $\text{bias}$ refers to the offset.

The gating branch directly sends the input $x_z$ through a linear projection layer followed by SiLU activation to produce its output. The output $y_u$ from the gating branch is then added element-wise to the output $y_t$ from the SSM branch to obtain an output that contains temporal context information. The resulting sequential features are passed into a fully connected layer for projection, yielding the final prediction results.

\subsection{Training Method}
In this paper, a novel loss function is proposed and the stage-wise training method from \cite{11,12} is adopted to train the model, aiming to improve overall performance.

This approach consists of two training stages, which are described in detail as follows.

(1) In the first stage, the output of FEB is combined with a fully connected layer to produce the results directly. The first two terms of the total loss function proposed in \cite{22} are used as the loss function to optimize the parameters.

\begin{equation}
\text{TotalLoss}_1 = (1-\lambda) \times \text{WCE} + \lambda \times \text{FL}
\label{eq:loss_stage1}
\end{equation}

Here, WCE refers to the weighted cross-entropy loss, and its calculation formula is as follows:

\begin{equation}
\text{WCE} = W \times \text{CE}(y, \hat{y})
\label{eq:wce}
\end{equation}

\begin{equation}
\text{CE} = -\frac{1}{N} \left\{ \sum_{n=1}^{N} \sum_{c=1}^{C} \left[ y_n^c \log(\hat{y}_n^c) + (1-y_n^c) \log(1-\hat{y}_n^c) \right] \right\}
\label{eq:ce}
\end{equation}

In this formula, CE represents the traditional cross-entropy loss. $y$ denotes the true label, and $\hat{y}$ represents the predicted class probability by the model. $N$ is the number of samples. $y_n^c$ and $\hat{y}_n^c$ are the true label and predicted probability of class $c$ for the $n$-th sample, respectively. $W$ represents the class weight.

FL refers to the dynamic focal loss, and its calculation formula is as follows:

\begin{equation}
\text{FL}(p_t, \alpha_t, \gamma) = -\alpha_t (1-p_t)^\gamma \log(p_t)
\label{eq:fl}
\end{equation}

$p_t$ represents the predicted probability by the model. $\alpha_t$ denotes the class weight, and $\gamma$ is the modulation factor. A larger value of $\gamma$ means the model pays less attention to easily classified samples and more attention to difficult samples. $t$ indicates the true label for each sleep stage.

(2) At the beginning of the second stage, the optimal FEB parameters obtained from the first stage are loaded and frozen, and only the parameters of the classification network are trained. The first two terms of the loss function are the same as in the first stage, while a third term, the contrastive loss (CL), is also introduced:

\begin{equation}
\text{TotalLoss}_2 = (1-\lambda) \times \text{WCE} + \lambda \times \text{FL} + \text{CL}
\label{eq:loss_stage2}
\end{equation}

\begin{figure}[!t]
\centering
\includegraphics[width=\linewidth]{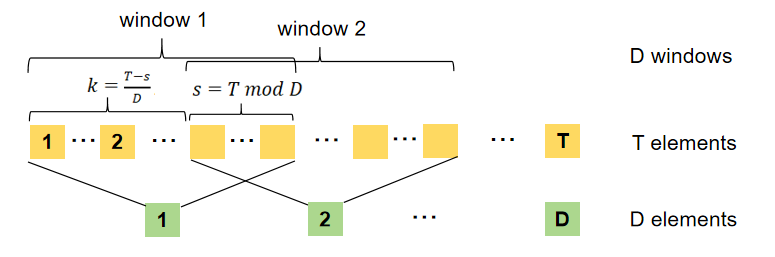}
\caption{Illustration of dynamic sliding window feature alignment for contrastive loss calculation.}
\label{fig:sliding_window}
\end{figure}

The CL loss is a contrastive loss between the outputs of the proposed model and the current SOTA model. The calculation method is illustrated in Fig.~\ref{fig:sliding_window}. Specifically, during the forward propagation in training, both the proposed model and the SOTA model are run to obtain the normalized features after temporal context capture, denoted as $Z_1 \in \mathbb{R}^{B \times L \times D}$ for the proposed model and $Z_2 \in \mathbb{R}^{B \times L \times T}$ for the SOTA model. Since the last feature dimension of the proposed model after forward propagation does not match that of the SOTA model, it is assumed that $T \geq D$.

To ensure that sequences of different lengths can be properly compared in the contrastive loss calculation, the sliding window parameters need to be dynamically computed according to the feature dimension. The total number of windows is fixed at $D$, the remainder $s = T \bmod D$, the window step size $k = \frac{T-s}{D}$, and the window size $w = s + k$. Based on this, the $i$-th feature window $W_{n,i} \in \mathbb{R}^{w}$ for the $n$-th sample corresponds to the $i$-th element of the short feature. The CL loss is defined as the mean squared error between the features within the $i$-th window and the $i$-th element of the short feature.
\section{Experiments Design}

\subsection{Dataset}
This study uses the Sleepedf (Sleepedf Expanded, 2018) dataset\cite{34,35} and a private dataset.

The private dataset was obtained from the First Hospital of Jilin University, with approval from the hospital's review board. The dataset includes overnight sleep recordings from 44 patients with sleep disorders, comprising 28 male and 16 female subjects. Data were collected using a PSG device and include eight-channel EEG recordings from the following electrode pairs: E1-M2, E2-M1, F3-M2, F4-M1, C3-M2, C4-M1, O1-M2, and O2-M1. Both EEG and EOG signals were sampled at a frequency of 256 Hz.

To investigate the shortest input window for real-time systems while ensuring performance, the initial EEG epoch length will be set to 30 seconds in accordance with previous research standards. When inputting data into the model, each 30-second epoch will be resampled to 100 Hz, and z-score normalization\cite{36} will be independently applied to each epoch. 

To explore the optimal dataset size, the dataset is processed into different sizes. The number of subjects and sample distributions for each dataset are shown in Table \ref{tab:dataset_stat}.

\begin{table}[htbp]
\centering
\caption{Experimental settings and dataset statistics.}
\label{tab:dataset_stat}
\begin{tabular*}{\linewidth}{l @{\extracolsep{\fill}} c c c c c c}
\hline
Dataset & \multicolumn{6}{c}{Class distribution} \\
& W & N1 & N2 & N3 & REM & Total \\
\hline
Sleepedf-20 & 4838 & 1689 & 8927 & 3074 & 3693 & 22221 \\
Sleepedf-25 & 6296 & 1950 & 11566 & 3452 & 4666 & 27930 \\
Sleepedf-30 & 7281 & 2225 & 13494 & 4152 & 5691 & 32843 \\
Sleepedf-35 & 9348 & 2434 & 15854 & 5026 & 6865 & 39527 \\
Sleepedf-40 & 10460 & 2843 & 18338 & 5707 & 7894 & 45242 \\
Sleepedf-45 & 11949 & 3467 & 20704 & 5995 & 8741 & 50856 \\
Private dataset & 10986 & 4849 & 16737 & 4206 & 6137 & 42915 \\
\hline
\end{tabular*}
\end{table}

\subsection{Experimental Setup}
The K-fold cross-validation method was used to evaluate the model's performance: for the Sleepedf dataset, K=10 (Fpz-Cz channel); for the private dataset, K=10 (C4-M1 channel).

At the beginning of each training session, 10 percent of the complete dataset is randomly selected as the test set. The remaining data are divided so that $N-\frac{N}{K}$ subjects are used as the training set, and $\frac{N}{K}$ subjects are used as the validation set, where $N$ is the total number of subjects in the remaining dataset.

Model performance is assessed using five metrics: overall accuracy (ACC), macro-averaged F1 score (MF1), Cohen's Kappa coefficient ($\kappa$), the number of model parameters, and the model computational cost.

\begin{equation}
\text{ACC} = \frac{\sum_{c=1}^{C} \text{TP}_c}{N}, \quad \text{MF1} = \frac{\sum_{c=1}^{C} \text{F1}_c}{C}
\label{eq:metrics}
\end{equation}

Here, $\text{TP}_c$ represents the number of true positive samples for class $c$, $\text{F1}_c$ is the F1 score for class $c$, $C$ is the number of sleep stages, and $N$ is the total number of test segments.

The parameters for each part of the model are shown in Table \ref{tab:model_params}.

\begin{table*}[htbp]
\centering
\caption{The parameters for each part of the model.}
\label{tab:model_params}
\begin{tabular}{lrlr}
\hline
LR & $5\text{e-}4$ & CE\_WEIGHT & 0.6 \\
WEIGHT\_DECAY & $1\text{e-}5$ & CONTRAST\_WEIGHT & 20.0 \\
PATIENCE & 50 & FEB\_FEAT\_DIM & 36 \\
CLASS\_WEIGHTS & $[1.0,2.3,1.1,1.3,1.7]$ & MAMBA\_N & 36 \\
FOCAL\_ALPHA & 0.25 & MAMBA\_CONV & 4 \\
FOCAL\_GAMMA & 2.0 & MAMBA\_DEPTH & 2 \\
\hline
\end{tabular}
\end{table*}

\subsection{Ablation Experiment Setup}
To verify the effectiveness of each part of the model, the following baseline models are implemented under a unified framework:
\begin{itemize}
    \item Traditional CNN with fully connected classifier (Baseline-1);
    \item Gabor kernel function replaces CNN with fully connected classifier (Baseline-2);
    \item FEB layer(Gabor kernels+learnable filters) with fully connected classifier (Baseline-3);
    \item GamSleepNet(FEB+Mamba) with fully connected classifier (Baseline-4);
\end{itemize}

All networks are trained and evaluated using the same dataset partitioning logic.

\section{Results}

\subsection{Optimal Dataset Size}
Most existing sleep staging models show a similar pattern during performance evaluation: When the dataset is too small, the model tends to overfit, resulting in poor generalization performance in the final model, as reported in previous studies\cite{8,9,11,16,18}. On the other hand, when the dataset is too large, the model often struggles to converge during training, leading to suboptimal final performance. In this study, models are trained using Sleepedf datasets of different sizes to explore the optimal dataset size. Detailed information about each dataset is provided in section 1 of the experimental design.

\begin{figure}[!t]
\centering
\includegraphics[width=\linewidth]{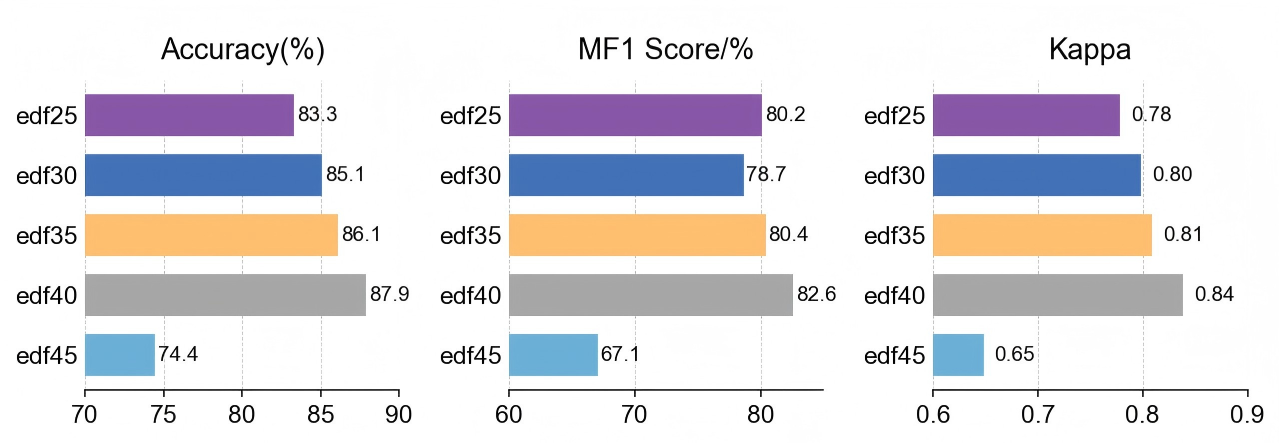}
\caption{Confusion matrices of models trained on datasets of different sizes}
\label{fig:fig3}
\end{figure}

The automatic sleep staging classification performance of models trained on datasets of different sizes is shown in Fig. \ref{fig:fig3}.

The results indicate that as the size of the Sleepedf dataset increases, the sleep staging performance of the GamSleepNet first shows a steady improvement and then a sharp decline. When the number of subjects rises from 25 to 40, the overall accuracy increases from 83.3\% to 87.9\%, the macro-average F1 score rises from 80.2\% to 82.6\%, and the Kappa coefficient climbs from 0.78 to 0.84. However, when the number of subjects reaches 45, all three key metrics drop significantly. The overall accuracy decreases to 74.4\%, the macro-average F1 score drops to 67.1\%, and the Kappa coefficient falls to 0.65. Further experiments show that as the dataset continues to grow, model performance declines more sharply and becomes much lower than at the edf45 level, so the details are not discussed further.

In summary, Sleepedf-40 represents the optimal training size for this model on the Sleepedf dataset, and all subsequent experiments are conducted based on this dataset. In addition, for practical deployment, the dataset size should be maintained between 30 and 40 subjects.

\subsection{Ablation Experiment}
To verify the effectiveness of GamSleepNet, ablation experiments are conducted in this section from two perspectives: baseline models and training methods. Section 1 evaluates the performance of the FEB+mamba architecture by replacing the five baseline models mentioned in the ablation experiment setup. Section 2 verifies the effectiveness of the training method by removing the contrastive loss during the second stage of training.

\subsubsection{Backbone Network Performance}
\label{sec:backbone_performance}
Table \ref{tab:backbone_performance} presents the performance of the five baseline models on the Sleepedf dataset.

\begin{table*}[htbp]
\centering
\caption{The performance of the four backbone networks on the Sleepedf dataset is shown here. Detailed descriptions of the four baselines are provided in section 1 of the experimental design.}
\label{tab:backbone_performance}
\begin{tabular*}{\linewidth}{l @{\extracolsep{\fill}} c c c c c c c c c}
\toprule
Model & ACC & MF1 & $\kappa$ & W & N1 & N2 & N3 & REM \\
\midrule
Baseline-1 & 79.65\% & 72.39\% & 0.7295 & 93.14\% & 42.41\% & 85.63\% & 79.46\% & 61.82\% \\
Baseline-2 & 79.53\% & 72.08\% & 0.7271 & 90.67\% & 41.95\% & 85.63\% & 82.08\% & 63.77\% \\
Baseline-3 & 84.72\% & 76.34\% & 0.7960 & 97.13\% & 27.93\% & 86.59\% & 91.16\% & 77.99\% \\
Baseline-4 & 87.86\% & 82.62\% & 0.8387 & 93.46\% & 54.07\% & 89.18\% & 89.42\% & 88.44\% \\
\bottomrule
\end{tabular*}
\end{table*}

According to the ablation results in Table \ref{tab:backbone_performance}, the model using only the traditional CNN architecture (Baseline-1) and the model replacing the convolution with Gabor filters (Baseline-2) show similar performance across all metrics. Their overall accuracy is below 80\%, macro F1-score is less than 73\%, and their ability to recognize minority classes is poor, with N1 at around 42\%  and REM at about 62\%. However, the performance for N3 improves from 79.46\% to 82.08\%. After introducing learnable filters based on Baseline-2 to form Baseline-3, the model performance increases significantly with accuracy rising to 84.72\%, macro F1-score reaching 76.34\%, and kappa increasing to 0.7960. The classification performance for the REM class also rises sharply to 77.99\%. On this basis, further introducing the mamba module to build the complete GamSleepNet (Baseline-4) leads to further improvements in overall metrics, with accuracy at 87.86\%, macro F1-score at 82.62\%, and kappa at 0.8387. The performance for difficult-to-classify samples improves substantially, with N1 increasing from less than 43\% to 54.07\%, and REM rising from about 78\% to 88.44\%. The classification performance for the other classes, W, N2, and N3, also remains at a high level.

In summary, from Baseline-1 to Baseline-4, as the FEB layer and the Mamba module are introduced sequentially, the overall accuracy, macro F1-score, and kappa of the model continue to improve. Compared with the first three baselines, the complete GamSleepNet achieves significantly better recognition of the hard-to-distinguish N1 and REM stages, demonstrating the effectiveness of the FEB and Mamba modules.

\subsubsection{Training Method Performance}
\label{sec:Training method performance}
Table \ref{tab:ablation_contrast_loss} and Table \ref{tab:contrast_loss_full} show the performance of GamSleepNet on the Sleepedf dataset without and with the proposed contrastive loss method, respectively.

\begin{table*}[htbp]
\centering
\setlength{\tabcolsep}{5pt}
\caption{The performance of GamSleepNet on the Sleepedf dataset without the proposed contrastive loss method.}
\label{tab:ablation_contrast_loss}
\begin{tabular*}{\linewidth}{c @{\extracolsep{\fill}} c c c c c c c c}
\toprule
Actual & \multicolumn{5}{c}{Predicted} & \multicolumn{3}{c}{Overall metrics} \\
& W & N1 & N2 & N3 & REM & ACC & MF1 & $\kappa$ \\
\midrule
W  & 94.54\% & 4.28\% & 0.63\% & 0.31\% & 0.23\% & \multirow{5}{*}{87.37\%} & \multirow{5}{*}{82.09\%} & \multirow{5}{*}{0.8324} \\
N1 & 19.98\% & 53.93\% & 10.86\% & 0.43\% & 14.79\% & & & \\
N2 & 0.17\% & 4.22\% & 88.43\% & 4.28\% & 2.90\% & & & \\
N3 & 0.01\% & 0.01\% & 10.86\% & 89.13\% & 0.00\% & & & \\
REM& 0.56\% & 7.28\% & 6.18\% & 0.00\% & 85.99\% & & & \\
\bottomrule
\end{tabular*}
\end{table*}

\begin{table*}[htbp]
\centering
\setlength{\tabcolsep}{5pt}
\caption{The performance of GamSleepNet on the sleepedf dataset with the proposed contrastive loss.}
\label{tab:contrast_loss_full}
\begin{tabular*}{\linewidth}{c @{\extracolsep{\fill}} c c c c c c c c}
\toprule
Actual & \multicolumn{5}{c}{Predicted} & \multicolumn{3}{c}{Overall metrics} \\
& W & N1 & N2 & N3 & REM & ACC & MF1 & $\kappa$ \\
\midrule
W  & 93.46\% & 5.53\% & 0.45\% & 0.29\% & 0.27\% & \multirow{5}{*}{87.86\%} & \multirow{5}{*}{82.62\%} & \multirow{5}{*}{0.8387} \\
N1 & 18.09\% & 54.07\% & 10.17\% & 1.03\% & 16.64\% & & & \\
N2 & 0.31\% & 3.33\% & 89.18\% & 3.40\% & 3.78\% & & & \\
N3 & 0.00\% & 0.00\% & 10.58\% & 89.42\% & 0.00\% & & & \\
REM& 0.56\% & 5.69\% & 5.31\% & 0.00\% & 88.44\% & & & \\
\bottomrule
\end{tabular*}
\end{table*}

Experimental results show that, in terms of overall performance, the model trained with the contrastive loss function proposed in this paper achieves improvements over the baseline model without this loss function. Specifically, the accuracy (ACC), macro F1 score (MF1), and Kappa coefficient increased by 0.49\%, 0.53\%, and 0.0063, respectively. At the category level, although the accuracy for the easily classified W stage slightly decreased by 1.08\%, the prediction accuracy for all difficult-to-classify categories improved significantly. The increases for the N1, N2, N3, and REM stages were 0.14\%, 0.75\%, 0.29\%, and 2.45\%, respectively. 

In summary, the loss function proposed in this paper can effectively enhance the model's recognition performance for difficult-to-classify categories.

\subsection{Comparison with SOTA Methods}
Table \ref{tab:sota_comparison} shows the performance of GamSleepNet and current SOTA models on the sleepedf dataset and private dataset. The evaluation metrics include confusion matrix, ACC, MF1, $\kappa$, and number of parameters.

\begin{figure}[htbp]
\centering
\includegraphics[width=0.5\textwidth]{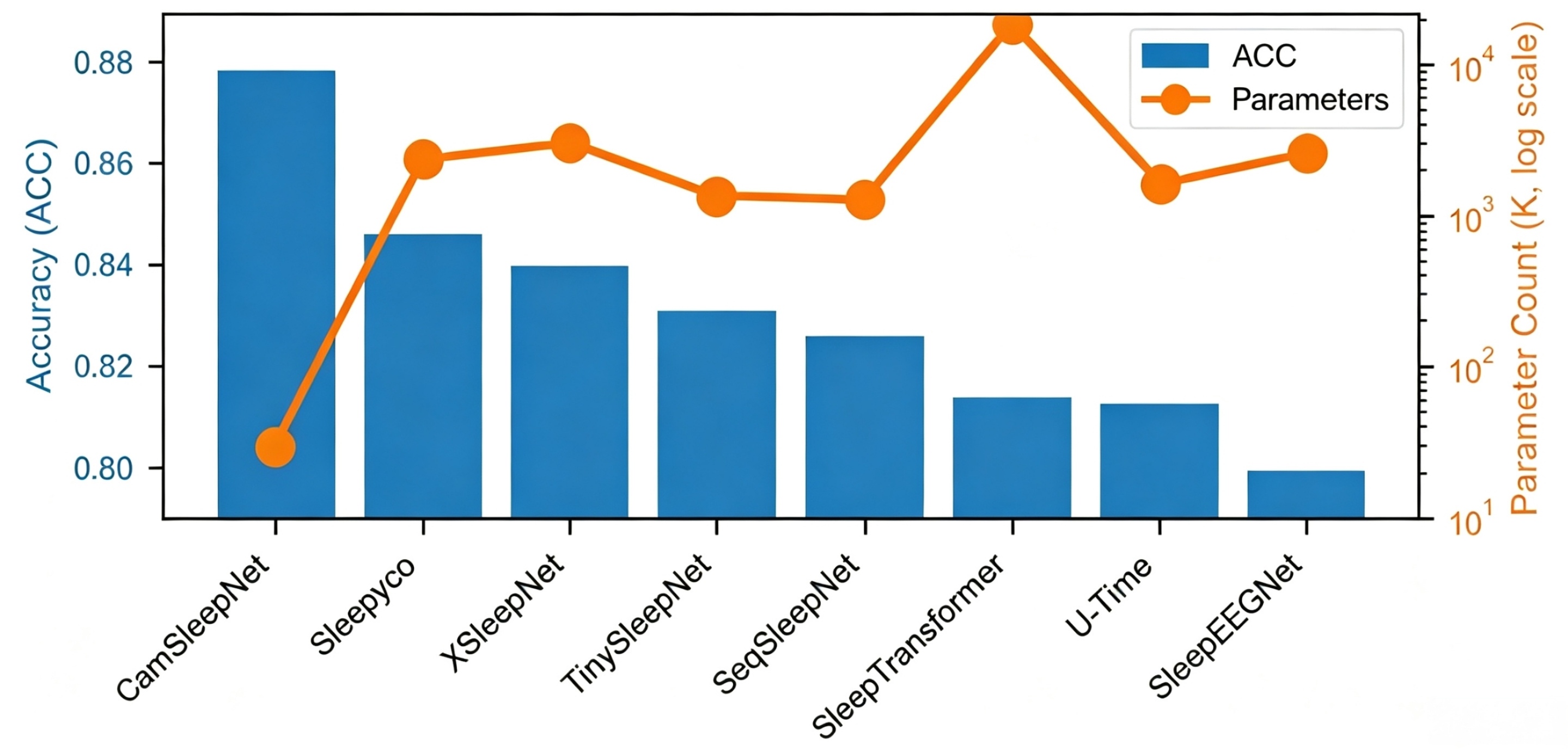}
\caption{Performance comparison of sleep staging models in terms of accuracy and parameter count.}
\label{fig:fig4}
\end{figure}

Figure \ref{fig:fig4} visually illustrates the performance comparison between GamSleepNet and the current state-of-the-art models in terms of accuracy (ACC) on sleepedf dataset and the number of parameters. 

\begin{table*}[htbp]
\centering
\setlength{\tabcolsep}{4pt}
\caption{The performance of GamSleepNet and current SOTA models on the sleepedf and private dataset.}
\label{tab:sota_comparison}
\begin{tabular*}{\linewidth}{l @{\extracolsep{\fill}} c c c c c c c c c c c}
\toprule
Method & Dataset & ACC & MF1 & $\kappa$ & W & N1 & N2 & N3 & REM & Parameters \\
\midrule
GamSleepNet (Ours) & \multirow{9}{*}{Sleepedf-SC} & 87.86 & 82.62 & 0.839 & 93.46 & 54.07 & 89.18 & 89.42 & 88.44 & 30.86k \\
Sleepyco\cite{11} & & 84.6 & 79.0 & 0.787 & 93.5 & 50.4 & 86.5 & 80.5 & 84.2 & 2.38M \\
Korkalainen et al.\cite{39} & & 83.7 & -- & 0.77 & -- & -- & -- & -- & -- & 1.53M \\
XSleepNet\cite{16} & & 84.0 & 77.9 & 0.778 & 93.3 & 49.9 & 86.0 & 78.7 & 81.8 & 3.00M \\
TinySleepNet\cite{40} & & 83.1 & 78.1 & 0.77 & 92.8 & 51.0 & 85.3 & 81.1 & 80.3 & 1.36M \\
SeqSleepNet\cite{7} & & 82.6 & 76.4 & 0.760 & -- & -- & -- & -- & -- & 1.28M \\
SleepTransformer\cite{10} & & 81.4 & 74.3 & 0.743 & 91.7 & 40.4 & 84.3 & 77.9 & 77.2 & 18.99M \\
U-Time\cite{8} & & 81.3 & 76.3 & 0.745 & 92.0 & 51.0 & 83.5 & 74.6 & 80.2 & 1.63M \\
SleepEEGNet\cite{41} & & 80.0 & 73.6 & 0.73 & 91.7 & 44.1 & 82.5 & 73.5 & 76.1 & 2.60M \\
\midrule
GamSleepNet (Ours) & \multirow{5}{*}{\shortstack{private\\dataset}} & 86.42 & 86.08 & 0.8199 & 83.58 & 78.09 & 81.54 & 93.31 & 95.00 & 30.86k \\
Sleepyco\cite{11} & & 86.13 & 81.05 & 0.7914 & 71.01 & 70.43 & 82.83 & 97.73 & 94.66 & 2.38M \\
SleepTransformer\cite{10} & & 85.18 & 79.74 & 0.7978 & 72.09 & 43.70 & 86.17 & 95.36 & 93.64 & 18.99M \\
DeepSleepNet\cite{12} & & 83.81 & 79.44 & 0.7591 & 89.51 & 64.21 & 77.89 & 98.11 & 95.02 & 24.7M \\
TinySleepNet\cite{40} & & 74.23 & 65.76 & 0.6116 & 77.35 & 60.00 & 80.32 & 80.46 & 54.11 & 1.36M \\
\bottomrule
\end{tabular*}
\end{table*}

The performance of GamSleepNet compared with SOTA frameworks is presented in Table \ref{tab:sota_comparison}, which includes overall classification metrics, per-stage classification accuracy, and model parameter counts for different automatic sleep staging models on the Sleepedf dataset. Figure \ref{fig:fig4} visually illustrates the performance comparison between GamSleepNet and current mainstream SOTA models in terms of accuracy and parameter count. As shown in Figure \ref{fig:fig4} and Table \ref{tab:sota_comparison}, GamSleepNet achieves exceptional lightweight design while delivering the best overall classification performance among all compared models, with each key metric significantly surpassing existing single-channel EEG sleep staging models.

At the overall performance level, GamSleepNet achieved an overall accuracy of 87.86\%, a macro-averaged F1 score (MF1) of 82.62\%, and a Cohen's Kappa coefficient ($Kappa$) of 0.839 on the Sleepedf dataset. Compared with the second-ranked Sleepyco\cite{11}, these three core metrics improved by 3.26\%, 3.62\%, and 0.052, respectively. On a private clinical dataset, GamSleepNet also maintained leading overall performance, with an overall accuracy of 86.42\%, a macro-averaged F1 score of 86.08\%, and a Kappa coefficient of 0.8199, demonstrating the model's generalization ability in real-world clinical scenarios.

In terms of classification performance for each sleep stage, GamSleepNet demonstrates significant improvement in the traditionally challenging N1 and REM sleep stages, while maintaining high accuracy in the easier-to-classify stages. On the Sleepedf public dataset, the accuracy for the N1 stage reaches 54.07\% and for the REM stage 88.44\%, representing maximum improvements of 3.07\% and 4.24\% over the compared SOTA models, respectively. The W, N2, and N3 stages also remain among the leading group of similar models. On the private clinical dataset, the model likewise exhibits superior performance on difficult-to-classify samples. Specifically, the accuracy for the N1 stage reaches 78.09\%, representing improvements of 7.66\% and 34.39\% over Sleepyco and SleepTransformer, respectively; the accuracy for the REM stage reaches 95.00\%, which is essentially on par with the current best level; the accuracy for the N3 stage reaches 93.31\%, indicating stable and reliable deep sleep stage recognition. Overall, the model is well-suited to meet the staging demands of real-world clinical scenarios.

In terms of model lightweight design and deployment adaptability, GamSleepNet demonstrates overwhelming advantages, with a parameter count of only 30.86 thousand, which is orders of magnitude smaller than all compared SOTA models. Specifically, its parameter count is only 1.3\% of Sleepyco at 2.38 million, 1.03\% of XSleepNet at 3.00 million, and 0.16\% of SleepTransformer at 18.99 million, which has the largest parameter count. As shown in Figure \ref{fig:fig4}, GamSleepNet achieves the highest classification accuracy while compressing the parameter count to only a fraction of existing SOTA models. Owing to the linear temporal modeling complexity of the Mamba architecture, GamSleepNet also offers extremely low inference latency, far surpassing traditional models based on CNN and Transformer architectures, and is perfectly suited for the computational constraints of portable sleep monitoring devices.

\section{Conclusion}
This paper presents the GamSleepNet model, which addresses two major challenges commonly found in current automatic sleep staging methods: large model sizes that lead to overfitting on small datasets and low accuracy in distinguishing between sleep stages. The proposed FEB module and the integration of the Mamba architecture enable lightweight and low-latency automatic sleep staging using single-channel EEG signals, offering a practical solution for algorithm deployment in portable sleep monitoring devices.

The experimental section first evaluates the optimal dataset size for training this model. The results indicate that the model performs best when trained on a dataset containing between 30 and 40 subjects, with 40 subjects identified as the optimal dataset size for training this model.

Ablation experiments show that the FEB module in the proposed GamSleepNet model significantly outperforms traditional CNN architectures in terms of parameter size and classification accuracy. After integrating the Mamba architecture for temporal feature extraction,   the classification performance for challenging samples such as N1 and REM is greatly improved. By combining GamSleepNet with the contrastive loss calculation method introduced in this paper, it is possible to leverage the strengths of multiple models without altering the model architecture, further enhancing the classification accuracy for N1 and rapid eye movement sleep stages.

This paper conducts comparative experiments on the publicly available Sleepedf dataset and private dataset, and the results demonstrate that the GamSleepNet model achieves SOTA performance. Furthermore, the design approach of this model can be extended to other sequence data classification tasks and is suitable for various scenarios involving time series signals with multi-scale features and high similarity between classes.

In future research, the model will be further optimized to support real-time sleep staging and multi-channel devices, making it even more suitable for portable equipment.

\section{Acknowledgment}
This work was supported by the Jilin Provincial Science and Technology Development Program (Project No. 20240304052SF).

Thanks to the KinaMind society for their inspiring environment and unwavering support.

\bibliography{reference}
\end{document}